# Performance Evaluation of Nonlinear Model Predictive and Sliding Mode Controllers on a 4-Cable Spatial CSPR in Simscape

Erfun Bayat Sarmadi*
Center of Advanced System and Technologies (CAST)
Department of Mechanical Engineering, University of Tehran
Tehran, Iran
bayat.sarmadi.er@ut.ac.ir

Ali Asadi*
Center of Advanced System and Technologies (CAST)
Department of Mechanical Engineering, University of Tehran
Tehran, Iran
Ali.Asadi.a@ut.ac.ir

Moosa Ayati
Department of Mechanical Engineering, University of Tehran
Tehran, Iran
m.ayati@ut.ac.ir

***Abstract*— This paper offers a comprehensive comparative study between Nonlinear Model Predictive Control (NMPC) and Sliding Mode Control (SMC) applied to a spatial Cable-Suspended Parallel Robot (CSPR) actuated by four cables. The robot is modeled in two parallel ways: (i) via an analytical dynamic model with included cable elasticity and frictional losses, and (ii) in MATLAB Simscape Multibody, capturing realistic physical effects. Both controllers are implemented under identical reference trajectories, system constraints, and tuning strategies. Our results show that NMPC achieves superior tracking accuracy and stronger disturbance rejection compared to SMC, but at the cost of higher computational demand. Moreover, both controllers were tested on an analytical model of the robot and a Simscape model. The close agreement between the two, for both control schemes, confirms the accuracy of the analytical formulation and its consistency with the physics-based simulation. The study highlights the value of physics-based simulation as a complementary tool for validating advanced controllers under realistic conditions, and demonstrates that NMPC offers superior performance for the studied high-precision, constrained CSPR.**

## I. Introduction

Cable-Suspended Parallel Robots (CSPRs) are attractive in robotics due to their light weight, large workspace, and potentially high accelerations. By suspending an end-effector via multiple cables anchored to fixed base points, CSPRs avoid heavy rigid arms and allow distributed actuation. However, they face considerable control challenges: the system is underactuated in some directions, cable tensions must always remain positive, the dynamics are strongly nonlinear, and practical effects like cable elasticity and friction further complicate control.

Traditional controllers such as PID struggle in such environments, particularly under fast maneuvers or disturbances. Sliding Mode Control (SMC) is often proposed because of its robustness to modeling errors and disturbances; nevertheless, its inherently discontinuous control action can lead to chattering and actuator wear. On the other hand, Nonlinear Model Predictive Control (NMPC) can optimize future trajectories while respecting constraints and better adherence to physical limitations.

While recent robotic control studies have explored integrating perception-driven or learning-based modules with classical controllers, such approaches often target simpler, fully actuated systems. For example, in [1], a fuzzy PID controller was combined with an Xception-based vision module for real-time face tracking on a 4-DOF manipulator. Although this demonstrates the growing trend of perception-integrated control, the underlying control law remains relatively straightforward and not suited for systems with strong nonlinear coupling or tension constraints. In contrast, CSPRs demand advanced nonlinear control strategies such as SMC and NMPC to manage their coupled dynamics and ensure stable tension distribution.

Recent works on robust sliding-mode strategies for cable-driven robots emphasize SMC's practical robustness but also document issues with chattering and implementation. Several experimental and simulation studies apply variants of SMC (including adaptive and finite-time versions) to suspended cable robots and demonstrate good disturbance rejection and tracking under modeling uncertainty, while acknowledging actuator wear and chattering as practical drawbacks. For example, experimental investigations comparing PID, PD, and sliding-mode controllers for a suspended CDPR show SMC's robustness in tracking tasks but also underline implementation concerns when moving from simulation to hardware [2-4].

In contrast, a growing body of literature applies Model Predictive Control (MPC) and nonlinear MPC specifically to cable-driven parallel robots. NMPC approaches are attractive because they can explicitly enforce cable-tension and actuator constraints and optimize performance over a prediction horizon, properties particularly useful for constrained, underactuated CSPRs. Recent studies report that NMPC (or closely related predictive schemes) improves path-tracking accuracy and constraint handling, and several works explicitly address cables with axial flexibility when designing the predictive controller. These NMPC results typically report improved adherence to bounds compared to conventional controllers, at the cost of increased online computation [5-7].

*Authors Contributed Equally.

Accurate modeling of cable elasticity, pulley kinematics, and frictional losses is critical for realistic simulation and controller validation, yet many earlier studies use simplified massless, inextensible cable assumptions. Papers that incorporate cable axial flexibility and friction (or examine pulley/cable kinematics) show that these nonidealities substantially affect tension distribution, platform stiffness, and transient response, factors that materially influence controller performance in practice. Work on frictional effects in cable systems and on elastic cable modeling, therefore, forms an important foundation for more realistic controller evaluation [8, 9].

Physics-based modeling environments such as MATLAB/Simscape Multibody are increasingly used to validate controllers because they can capture multi-body interactions, pulley geometry, contact, and nonideal damping effects that are tedious to include analytically. Several recent papers and application notes show Simscape models of cable robots (including studies that explicitly compare Simscape and analytical results), arguing that Simscape is a useful complementary validation tool even when an analytical model is available. These studies motivate dual-implementation evaluation (analytic + Simscape) when the goal is to assess real-world viability rather than just theoretical feasibility [10].

Although prior literature includes studies on MPC or SMC for cable-driven robots, very few have (1) modeled cable elasticity and frictional losses explicitly, (2) implemented both controllers on an identical platform, and (3) validated them in a physics-based simulation environment such as Simscape. In this work, we close that gap. We develop a comparative framework in which both NMPC and SMC operate on the same robot dynamics, and in both analytic and Simscape domains. The main contributions of this study are as follows:

1) **Coupled Modeling of Elastic Cables and Friction:** We include cable elasticity and frictional losses in our analytic model, addressing realism that many prior works neglect.

2) **Side-by-Side Controller Design and Fair Comparison:** Both NMPC and SMC are tuned under identical trajectories, constraints, and actuator limits, enabling objective performance comparison.

3) **Dual Implementation: Analytic vs. Simscape:** Controllers are evaluated on both the analytical and Simscape models, confirming that the analytical model faithfully captures the system dynamics.

4) **Comparison Across Metrics:** The controllers are compared in terms of key performance aspects, including tracking accuracy, control effort, and robustness, for both analytical and Simscape implementations.

5) **Practical Insights for CSPR Designers:** The results guide future system designers in choosing control strategies and emphasize the importance of realistic simulation in validation.

The remainder of this paper details the modeling, controller design, and comparative simulations that support these contributions.

## II. System Modeling

### A. System Modeling with Elasticity and Friction of CSPR

We consider the 4-cable spatial CSPR shown in Figure 1. Let fixed anchor points be $A_i \in \mathbb{R}^3$, $\quad i = 1, \dots, 4$. The end-effector (EE) has attachment points in its body frame. The EE position in the global inertial frame is $p = [x, y, z]^T$. The vector from anchor to attachment is:

$$d_i = A_i - (p + B_i), \quad l_{i,0} = \|d_i\| \tag{1}$$

We adopt the unit cable direction:

$$\hat{\imath} = \frac{d_i}{\|d_i\|} \tag{2}$$

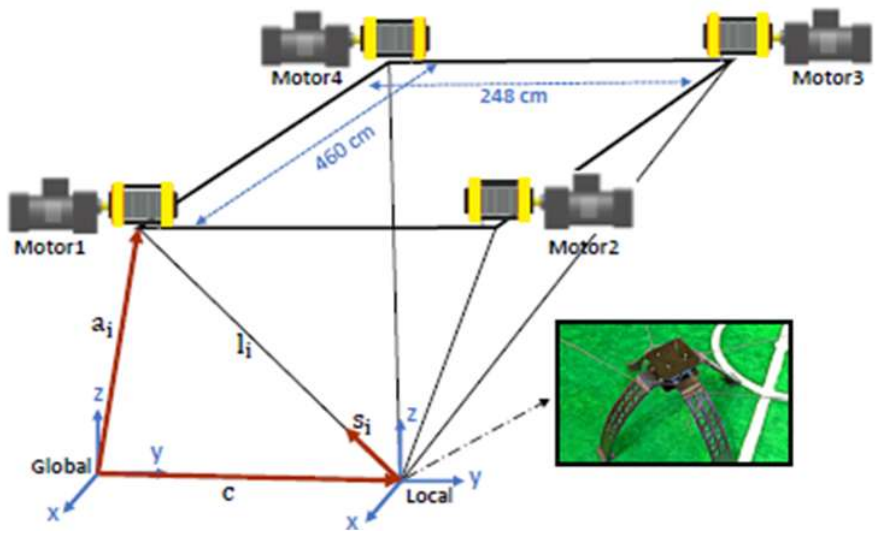

Figure 1. A schematic diagram of the CSPR under study [7]

Taking the derivative:

$$\dot{l}_i = -s_i^T \dot{c} \tag{3}$$

The Jacobian relating EE velocity $\dot{p}$ to the rate of cable length change $\dot{l}_i$ is:

$$\dot{l}_i = -\hat{u}_i^T \dot{p}, \quad \text{so} \quad J = \begin{bmatrix} -\hat{u}_1^T \\ -\hat{u}_2^T \\ -\hat{u}_3^T \\ -\hat{u}_4^T \end{bmatrix} \tag{4}$$

Thus $\dot{l} = J\dot{p}$.

### B. Dynamic Model Using Newton-Euler

To derive the full dynamics, we model both the winch subsystems and the EE. Winch Dynamics is as follows:

$$I\ddot{\theta}_i + F_v\dot{\theta}_i - RT_i = \tau_i \tag{5}$$

- $I$: Inertia of the winch
- $F_v$: Viscous friction coefficient
- $R$: Winch radius
- $T_i$: Tension in the cable $i$
- $\tau_i$: Applied torque to the winch $i$

End-Effector Dynamics, Using Newton's second law in 3D space:

$$\sum T_i\, s_i - mg\hat{z} = m\ddot{c} \tag{6}$$

In matrix notation, we have:

$$J^T F - G = M\ddot{x} \tag{7}$$

Where:

- $J^T F$: Total force from cables

- $G = [0,0,mg]^T$: Gravity vector
- $M$: EE mass

Combining with winch dynamics, the complete robot model becomes:

$$M\ddot{x} + G = -J^T R^{-1}(I\ddot{\theta} + F_v\dot{\theta} - \tau) \tag{8}$$

Substituting joint dynamics into Cartesian space via:

$$\dot{\theta} = R^{-1}J\dot{x}, \quad \ddot{\theta} = R^{-1}(J\ddot{x} + \dot{J}\dot{x}) \tag{9}$$

Leads to the final form:

$$M_T\ddot{x} + C_T\dot{x} + G = J^T R^{-1}\tau \tag{10}$$

Where:

$$M_T = M + J^T R^{-1} I R^{-1} J \tag{11}$$

$$C_T = J^T R^{-1}(I R^{-1}\dot{J} + F_v R^{-1} J) \tag{12}$$

Table 1. Robot Physical Parameters

| Parameter | Symbol | Value | Unit |
|---|---|---|---|
| Mass of End-Effector | ( M ) | 3.05 | kg |
| Winch Radius | ( R ) | 3.79 | cm |
| Winch Inertia | ( I ) | 0.00219 | kg·m² |
| Friction Coefficient | ( F_v ) | 0.1 | N·m·s |
| Gravitational Acceleration | ( g ) | 9.81 | m/s² |
| Cable Count | — | 4 | — |

### C. Cable Elasticity and Friction Modeling

To reflect realistic behavior, each cable is modeled as an elastic (linear spring) element plus viscous friction:

$$T_i = k_i(l_i - l_{i,0}) + c_i\dot{l}_i \tag{13}$$

Where $k_i$ is the stiffness, and $c_i$ is the damping coefficient of cable $i$. This captures elongation under tension and energy dissipation.

Furthermore, friction in the winch-pulley system is modeled as viscous friction at the motor:

$$\tau_{\text{fric},i} = f_{v,i}\dot{\theta}_i \tag{14}$$

Where $\dot{\theta}_i = \frac{\dot{l}_i}{R_i}$ Combined, the torque input equation becomes:

$$\tau_i = R_i T_i + I_i\ddot{\theta}_i + \tau_{\text{fric},i} \tag{15}$$

## III. Simscape Implementation and Model Verification

To bridge the gap between theoretical modeling and practical control deployment, the CSPR is reconstructed using MATLAB's Simscape Multibody environment. This modeling approach ensures that all physical interactions, such as gravitational forces, inertial effects, cable dynamics, and winch torque transmission, are computed in a high-fidelity physics engine.

The Simscape model shown in Figure 2, is used to validate the analytical dynamic model by applying identical input torques in [7]. Motor angle trajectories and end-effector displacements are compared. The normalized RMSE across all motors remained below 3.1% and all joints showed NRMSD below 0.5%, confirming excellent agreement between the analytical model and Simscape.

Importantly, evaluating controllers in Simulink allows for the realistic simulation of actuator limits, multibody dynamics, and nonideal effects, providing confidence that analytical-model-based controllers will perform as expected in practical implementation.

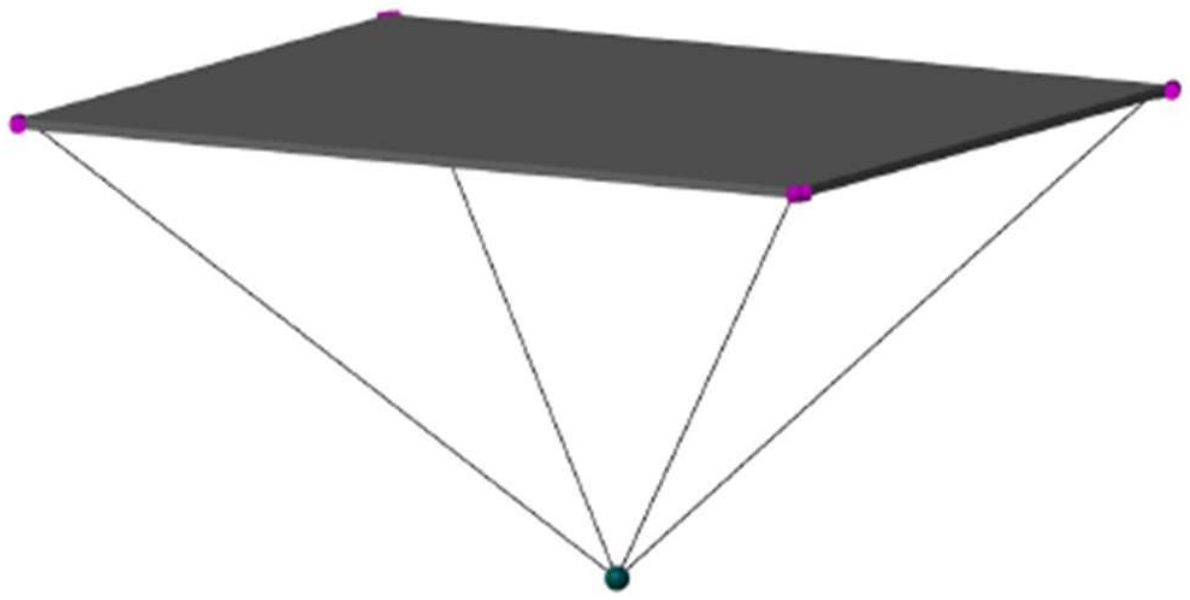

Figure 2. The physical shape of the CSPR modeled in Simscape [7]

## IV. Control Strategies

### A. Nonlinear Model Predictive Control (NMPC)

NMPC uses the discretized nonlinear dynamics of the CSPR to predict future states over a finite horizon. At each time step, it solves an optimization problem to minimize a cost function while satisfying state and input constraints [11]. The cost function used in this problem is as follows:

$$min \sum_{j=1}^{p} \| \hat{x}^{(k+j|k)} - w(k+j)\|^2 + \lambda \sum_{i=0}^{m-1} \| \tau(k+i)\|^2 \tag{16}$$

Subject to Constraints:

- Discretized dynamics (Euler method)
- Cable tension limits: $T_k \in [T_{min}, T_{max}]$
- Motor torque bounds: $\tau_k \in [\tau_{min}, \tau_{max}]$
- EE position bounds: $\hat{x}_k \in [\hat{x}_{min}, \hat{x}_{max}]$
- $k = 1, 2, \cdots, m$

The NMPC controller output is determined by solving the optimization problem with MATLAB's built-in function fmincon from the Optimization Toolbox. The controller is configured with:

- Prediction horizon: $p = 3$ steps
- Control horizon: $m = 3$ steps
- Sampling time: 0.01 s

### B. Sliding Mode Control (SMC)

SMC enforces system stability by driving the system state to a predefined sliding surface [12]. The sliding surface is defined as:

$$s = \dot{e} + \lambda e \tag{17}$$

Where $e = P_d - P$ is the position tracking error, and λ is a positive gain. The control law is derived using Lyapunov theory:

$$\tau = J^T[M(\ddot{p}_d + \lambda\dot{e}) + Mg\hat{z} + K \cdot \text{sat}\left(\frac{s}{\phi}\right)] \tag{18}$$

Here, $K$ is the control gain, and $\phi$ is the boundary layer to reduce chattering. Although SMC provides robust performance against uncertainties, it suffers from high-frequency torque oscillations, especially near the sliding surface. Overall, SMC provides strong robustness against unmodeled dynamics and disturbances, but may suffer from chattering if not carefully tuned.

## V. Simulation and Results

All simulations were performed using MATLAB Simulink and Simscape Multibody. Identical reference trajectories, initial conditions, physical parameters, and control bounds are used for both control strategies to ensure a fair comparison. The results are divided into three parts: trajectory tracking, control input profiles, and quantitative error analysis.

$$\begin{cases} x_r(t) = 0.6\cos(0.35t) \\ y_r(t) = 0.4\sin(0.35t) \\ z_r(t) = 0.2\cos\left(0.35t - \frac{\pi}{4}\right) \end{cases}, \quad t \in [0,100] \tag{19}$$

### A. Performance Metrics

The performance of both controllers is evaluated using several key metrics that capture tracking precision, control quality, robustness, and computational efficiency.

- **Root Mean Square Error (RMSE):** Average deviation between the reference and actual end-effector (EE) positions, indicating overall tracking accuracy.
- **Peak Error:** Maximum instantaneous deviation from the reference trajectory, reflecting transient performance.
- **Control Smoothness:** Degree of continuity in the generated torque inputs, related to actuator wear and mechanical stability.
- **Sensitivity to Parameter Mismatch:** Controller response to variations in system parameters such as cable stiffness and damping.
- **Computation Time per Control Step:** Average time required for each control iteration, indicating real-time feasibility.

### B. Results Overview

Figure 3 and Table 2 summarize the comparative performance of NMPC and SMC applied to both the analytical and Simscape models. NMPC achieves a lower average RMSE (1.11 mm) compared to SMC (2.04 mm), indicating higher tracking precision. However, SMC exhibits a slightly smaller maximum error (2.9 mm vs. 3.5 mm). The NMPC controller produces bounded torque inputs, whereas SMC shows discontinuities and mild chattering. In terms of computational demand, NMPC requires higher processing time per step, while SMC performs faster. Both controllers demonstrate robustness to noise and modeling errors, with SMC showing marginally higher resilience, and NMPC offering superior accuracy.

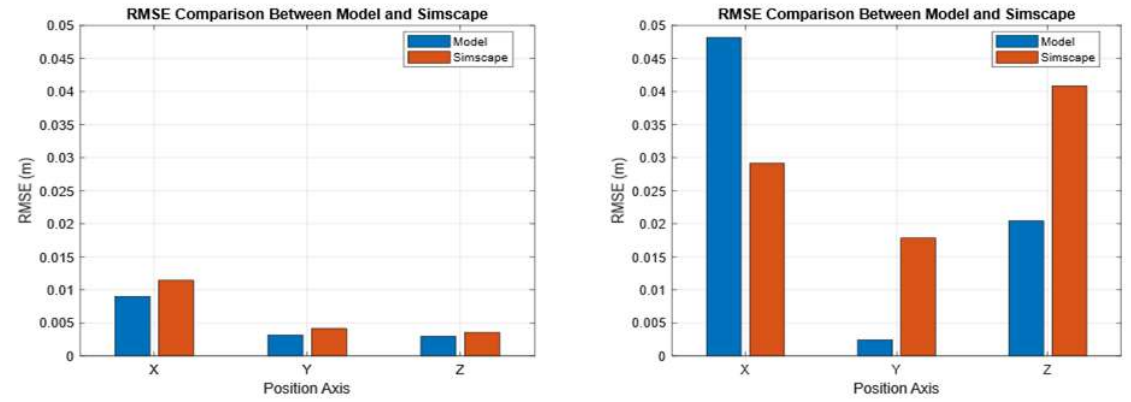

a) NMPC b) SMC

Figure 3. RMSE comparison between the Simscape model and analytical model using the NMPC & SMC controller

Table 2. Table of results and error

| Metric | MPC | SMC |
|---|---|---|
| Average RMSE | **5.72 mm** | 26.5 mm |
| Max tracking error | **8.59 mm** | 685.2 mm |
| Control smoothness | Sharp | **Smooth for Simscape,** Sharp for analytical model |
| Computation time | High | **Moderate** |
| Robustness to noise | Moderate | **High** |
| Robustness to modeling error | Moderate | **High** |

### C. Trajectory Tracking Plots

1) Position Tracking:

Figure 4-Figure 7 illustrate the end-effector (EE) trajectory tracking performance of both controllers. As shown in Figure 4 and Figure 5, the NMPC controller closely follows the reference trajectory with a lower average error and in all three directions (X, Y, and Z). In contrast, Figure 6 and Figure 7 show that the SMC controller also achieves successful tracking but exhibits occasional errors.

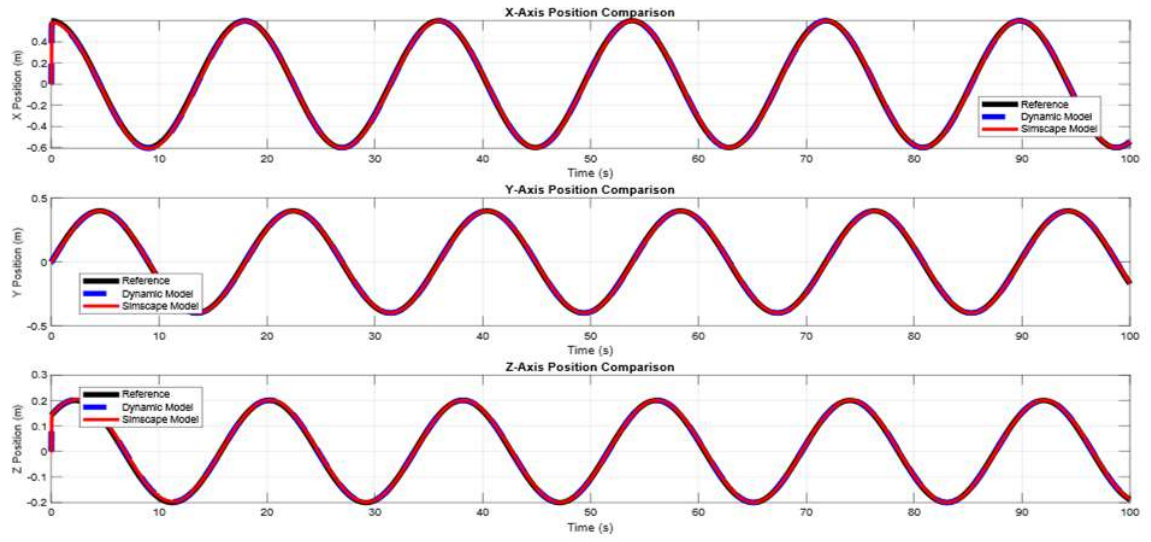

Figure 4. EE trajectory tracking using the NMPC controller compared to the reference

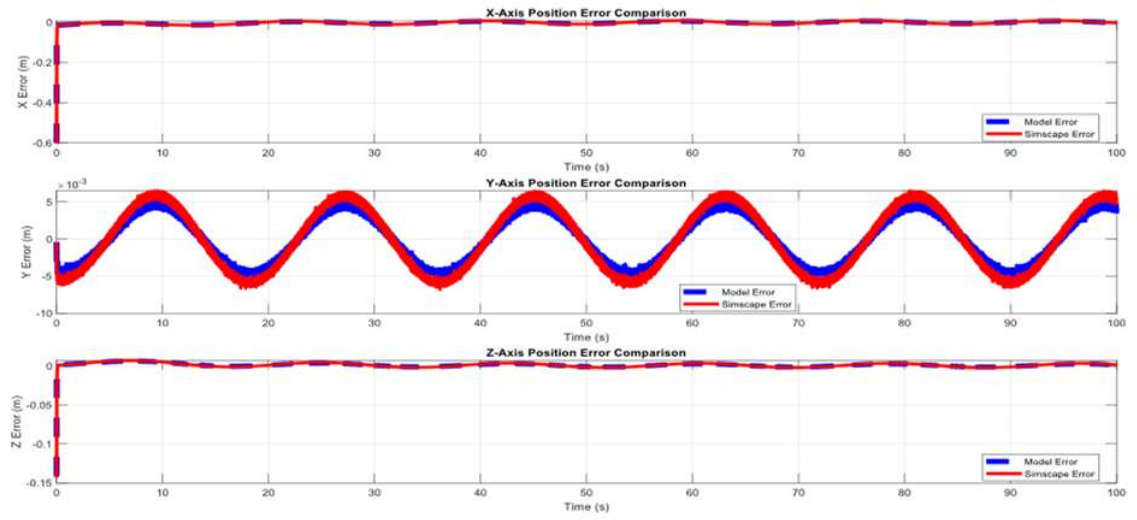

Figure 5. Tracking error in X, Y, and Z directions under NMPC control

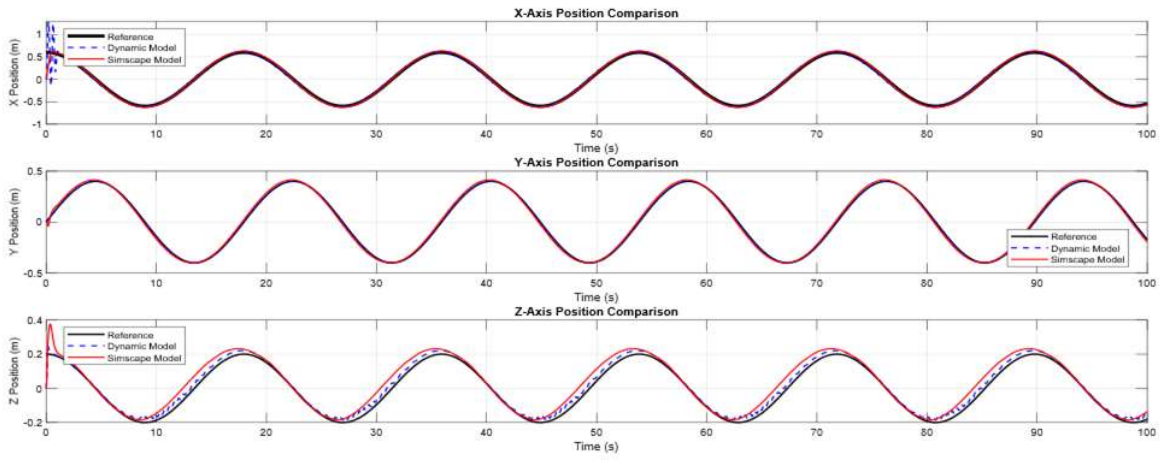

Figure 6. EE trajectory tracking using the SMC controller compared to the reference

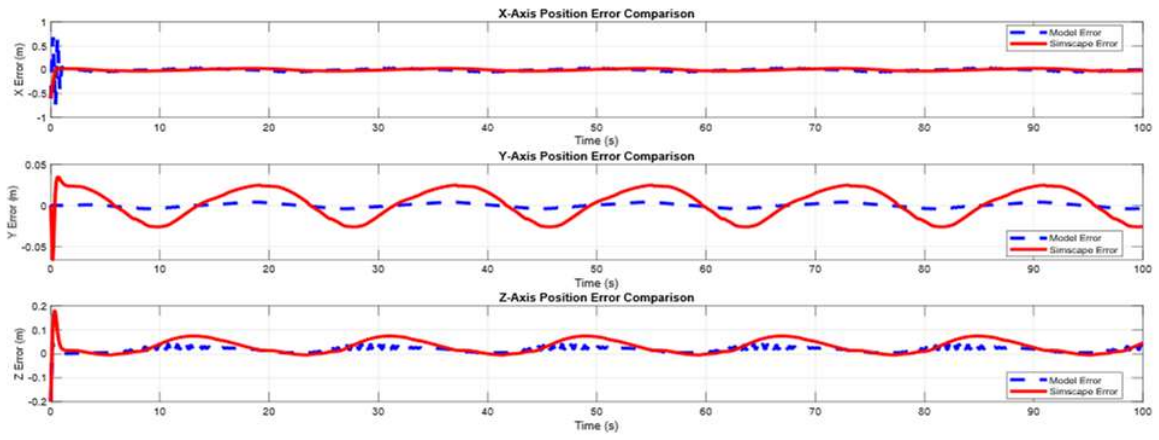

Figure 7. Tracking error in X, Y, and Z directions under SMC control

Figure 8 and Figure 9 present the three-dimensional end-effector (EE) trajectories for NMPC and SMC, respectively, compared to the reference path. The NMPC controller produces a trajectory that remains closely aligned with the desired path throughout the workspace, showing minimal deviation. In contrast, the SMC trajectory follows the general path shape but with slightly larger fluctuations near direction changes, consistent with its sharper control action observed in the time-domain plots.

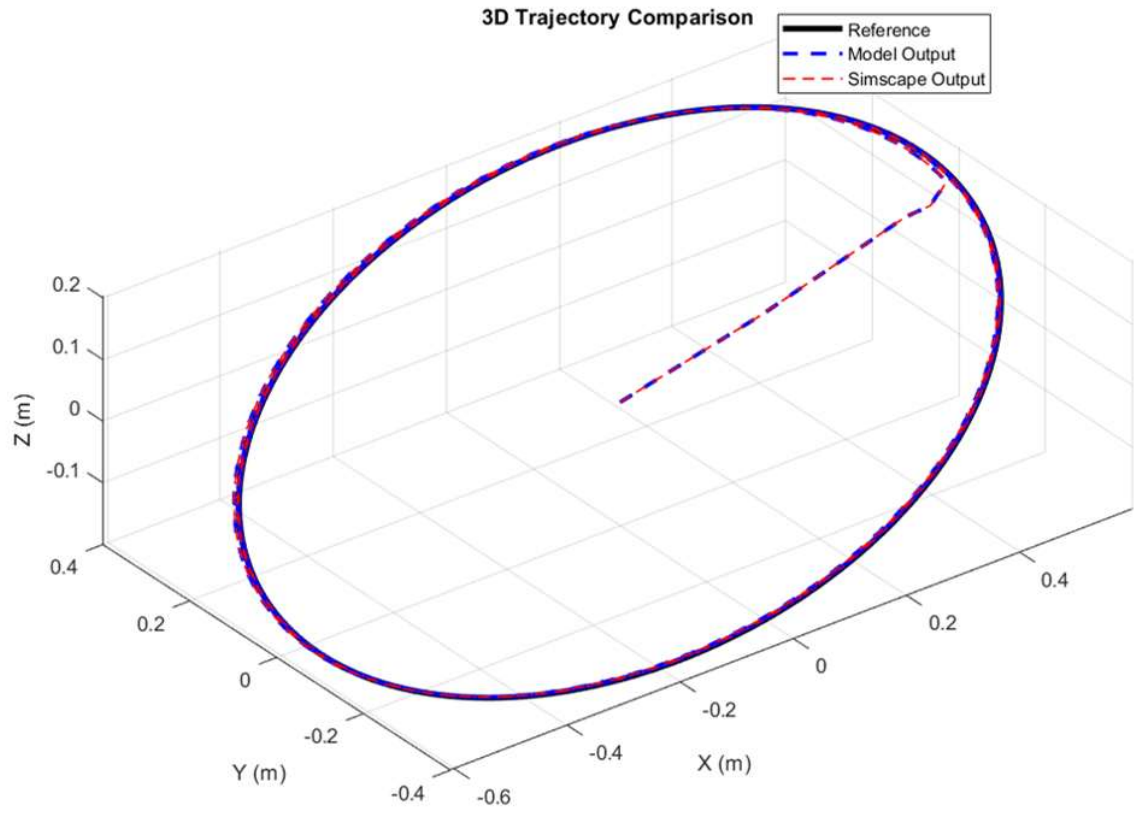

Figure 8. 3D end-effector trajectory using NMPC controller compared to the reference trajectory

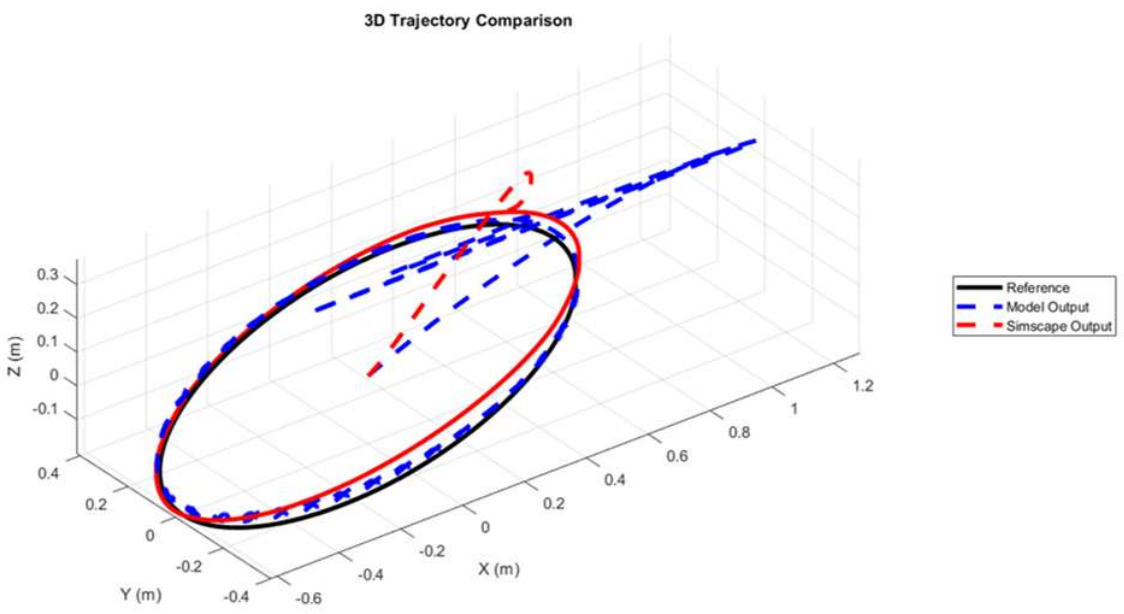

Figure 9. 3D end-effector trajectory using SMC controller compared to the reference trajectory

2) Torque Signals

Figure 10 and Figure 11 compare the actuator torque profiles generated by the two controllers for both modeling environments. The NMPC controller produces relatively sharp input variations, particularly around direction changes, indicating aggressive control action that could lead to mechanical stress in real CSPR hardware. In contrast, SMC yields discontinuous and chattering-like signals when applied to the analytical model but generates noticeably smoother torques in the Simscape environment. This difference suggests that physics-based modeling can mitigate some of SMC's discontinuities and that, in practice, SMC may provide a safer and more hardware-friendly behavior than NMPC, despite its lower tracking precision.

Figure 12 and Figure 13 illustrate the motor angle variations under NMPC and SMC control, respectively. Both controllers maintain stable and bounded motor motion throughout the trajectory.

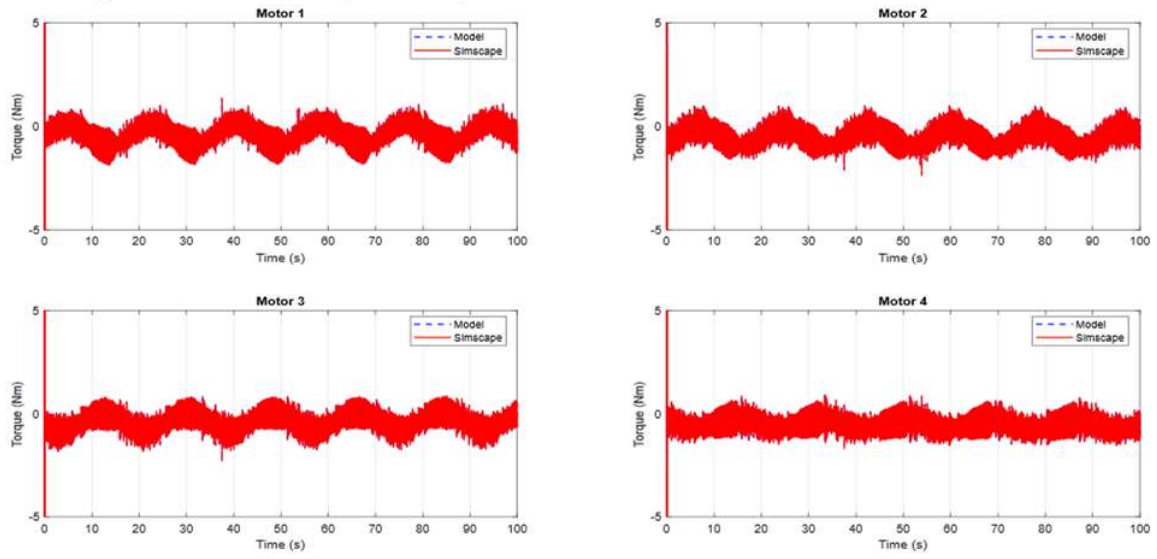

Figure 10. Control input torques generated by the NMPC controller

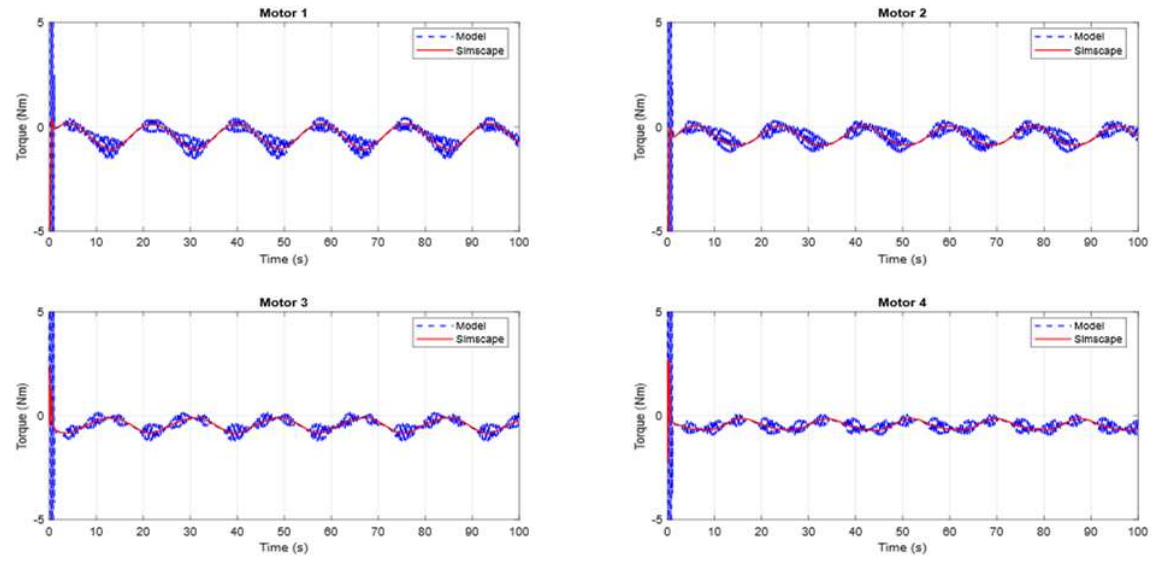

Figure 11. Control input torques generated by the SMC controller

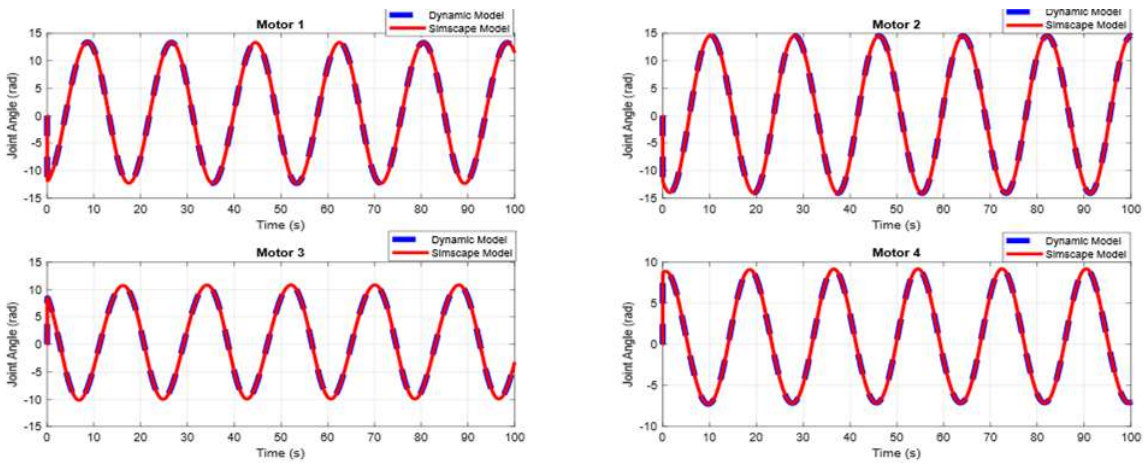

Figure 12. Motor Angles in case of using NMPC

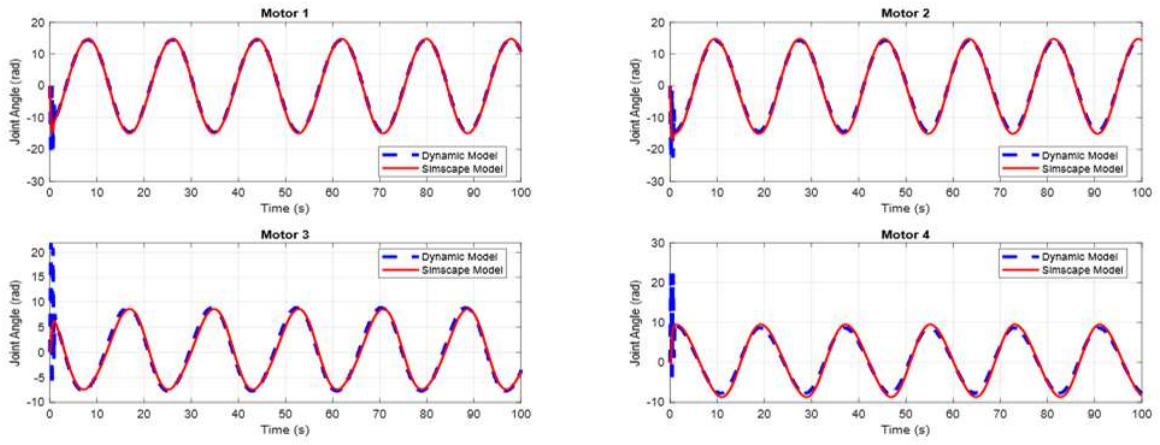

Figure 13. Motor Angles in case of using SMC

3) Robustness Test

In a robustness test, we introduce a 10% mass perturbation. MPC's performance degrades noticeably with increased RMSE up to 20.2 mm, while SMC maintains acceptable performance with RMSE under 30 mm due to its structure-invariant nature.

## VI. Discussion

### A. *Why NMPC Outperforms SMC in tracking?*

The superior tracking performance of NMPC arises from its predictive nature. By optimizing control actions over a finite horizon, it anticipates future reference changes and system constraints, enabling proactive, well-balanced inputs that enhance accuracy and stability. In contrast, SMC reacts to instantaneous errors, which can cause overcorrection or delayed convergence during rapid or coupled motions, leading to higher tracking errors.

### B. *Robustness and Adaptability*

Although less precise in tracking, SMC offers stronger robustness to parameter variations and unmodeled dynamics, maintaining stability under disturbances or parameter drift without relying on exact modeling. In contrast, NMPC performs better under nominal conditions but is more sensitive to model mismatch and requires careful tuning for robustness.

### C. *Role of Simscape in Enhancing Performance Insights*

Using MATLAB Simscape bridges the gap between theory and practice by incorporating effects like cable elasticity, friction, and multibody coupling, enabling controllers to respond more realistically to nonidealities. This realism helps assess practical feasibility and identify implementation issues before hardware testing, making Simscape a valuable intermediate step in controller validation.

### D. *Practical Trade-offs*

Comparing computational and practical aspects highlights clear trade-offs. NMPC’s online optimization imposes a heavy computational burden and depends strongly on model accuracy, making it sensitive to unmodeled disturbances. SMC, by contrast, runs efficiently in real time but still exhibits residual chattering despite boundary-layer mitigation. As the results are simulation-based, experimental validation is needed to confirm real-world performance.

### E. *Guidelines for Practitioners*

For practical use, controller choice depends on priorities and resources. NMPC suits applications demanding high accuracy and constraint handling when computation allows, while SMC is preferable for its simplicity, robustness, and real-time efficiency. In all cases, controllers should be co-tuned and validated in a physics-based environment like Simscape to ensure reliability under real-world conditions.

## VII. Conclusion

This study compared Nonlinear Model Predictive Control (NMPC) and Sliding Mode Control (SMC) on a four-cable spatial CSPR, incorporating cable elasticity and friction in both analytical and Simscape models. Results show that NMPC achieves higher tracking accuracy and disturbance rejection at the expense of greater computational demand. The strong agreement between analytical and Simscape results validates the analytical model, while Simscape offers a more realistic assessment by capturing physical nonidealities. Overall, the work underscores the value of physics-based modeling for reliable controller validation and tuning before hardware implementation.

For future work, the controllers will be implemented on a prototype CSPR platform to experimentally validate the simulation results. Further research will focus on developing computationally efficient NMPC variants and exploring adaptive or learning-enhanced extensions of both NMPC and SMC to handle time-varying parameters and improve robustness.